\documentclass[fleqn,10pt]{wlscirep}
\usepackage[utf8]{inputenc}
\usepackage[T1]{fontenc}
\usepackage{lineno}
\title{Cross-cultural Deployment of Autonomous Vehicles Using Data-light Inverse Reinforcement Learning}

\author[1,2,$\dag$,*]{Hongliang Lu}
\author[1,$\dag$]{Shuqi Shen}
\author[1]{Junjie Yang}
\author[3]{Chao Lu}
\author[1,*]{Xinhu Zheng}
\author[1,2,*]{Hai Yang}

\affil[1]{The Hong Kong University of Science and Technology (Guangzhou), Systems Hub}
\affil[2]{The Hong Kong University of Science and Technology, Civil and Environmental Engineering}
\affil[3]{Beijing Institute of Technology, Mechanical Engineering}

\affil[*]{corresponding authors: Hongliang Lu, Xinhu Zheng, and Hai Yang (hlu592@connect.hkust-gz.edu.cn; xinhuzheng@hkust-gz.edu.cn; cehyang@ust.hk)}

\affil[$\dag$]{these authors contributed equally to this work}

\begin{abstract}

More than the adherence to specific traffic regulations, \textbf{driving culture} touches upon a more implicit part - an informal, conventional, collective behavioral pattern followed by drivers - that varies across countries, regions, and even cities.
Such cultural divergence has become one of the biggest challenges in deploying autonomous vehicles (AVs) across diverse regions today. The current emergence of data-driven methods has shown a potential solution to enable culture-compatible driving through learning from data, but \textbf{what if some underdeveloped regions cannot provide sufficient local data to inform driving culture?} This issue is particularly significant for a broader global AV market.
Here, we propose a \textbf{cross-cultural deployment scheme for AVs}, called data-light inverse reinforcement learning, designed to re-calibrate culture-specific AVs and assimilate them into other cultures.
First, we report the divergence in driving cultures through a comprehensive comparative analysis of naturalistic driving datasets on highways from three countries: Germany, China, and the USA.
Then, we demonstrate the effectiveness of our scheme by testing the expeditious cross-cultural deployment across these three countries, with cumulative testing mileage of over
56,084 km.
The performance is particularly advantageous when cross-cultural deployment is carried out without affluent local data. Results show that we can reduce the dependence on local data by a margin of 98.67\% at best.
This study is expected to bring a broader, fairer AV global market, particularly in those regions that lack enough local data to develop culture-compatible AVs.
\\


\end{abstract}

\begin{document}

\flushbottom
\maketitle
%  Click the title above to edit the author's information and abstract

\thispagestyle{empty}

\section*{Introduction}\label{sec1}

%% 第一段
% 
What is \textbf{driving culture}? From the perspective of social behavior,
human behavior, as a consequence of socialization, is largely shaped by the sociocultural context in which individuals are immersed \cite{schill2019more,hodel2024characterizing}. As such, 
%although sometimes devoid of explicit and easy-to-access physical beings to capture and formalize culture, 
social behavior furnishes a substantial, versatile framework for accounting for how groups identify themselves and interact with the external world \cite{moeckli2007making,koster2020life}.
Indeed, human driving can be absorbed into this framework as driving behavior is inherently a social behavior characterized by moment-to-moment interactions among road users \cite{wang2022social}.
While driving, human drivers always follow a collective behavioral pattern acknowledged in their local driving community, oftentimes even without conscious awareness \cite{riener201617, faria2020assessing}. 
Such a pattern takes the form of an unspoken consensus conformed by drivers in their interactions with one another, existing beyond specific traffic regulations and sometimes even against them \cite{tennant2021code,redshaw2006driving}.
Accordingly, we define driving culture through the lens of driving behavior: 

\begin{quote}
    “\textit{More than mere adherence to specific traffic regulations, driving culture is an informal, conventional, collective behavioral pattern exhibited by drivers within their local driving community}.”
\end{quote}
For example, courteous driving is strongly embraced in some regions, whereas it may be quite the opposite in others; also, drivers in certain regions tend to adhere more strictly to traffic regulations, while in other regions drivers may show less respect for these regulations.
Such divergence, in a well-characterized yet intractable fashion, persists to varying degrees across different regions, cities, and even countries, thus contributing to the diversity and plenitude of driving culture today \cite{zaidel1992modeling}.
Yet, it is within this diverse and intricate landscape of driving cultures that the global deployment of autonomous vehicles (AVs) is facing tricky challenges.

%% 第二段
% 
At present, the pursuit of AV development is universally accepted as a vital global undertaking, which is largely expected to exert a profound positive impact on long-standing transportation systems \cite{bonnefon2016social,abdel2024matched}.
However, despite sounding promising, the widespread deployment of AVs remains challenging worldwide, as these newcomers, to date, have continued to adhere solely to pre-established strategies and fall short of absorbing and assimilating local driving cultures into their operations \cite{rahwan2019machine}. 
This perceived incongruity between AVs and human drivers is eroding public trust \cite{lee2004trust}, let alone achieving their seamless integration into existing transportation systems. 
More fundamentally, as previously mentioned, the intrinsic characteristics of culture render it impervious to effective handling through long-relied-on hard coding \cite{censi2019liability}.
To address this, recent years have witnessed a spurt of progress in capitalizing on data-driven approaches to capture driving culture from naturalistic driving data, thanks to the prowess of machine learning. %Some society-centered considerations have emerged accordingly.
For example, some scholars calibrated AVs by integrating the interactive strategies learned from data \cite{schwarting2019social,wang2021socially}; many studies proposed to develop human-like driving to embed humanness into AVs \cite{kolekar2020human,zhu2018human,lu2023human}.
Although these data-driven practices showcase a series of performant society-centered solutions, the biggest hurdle remains unsolved: \textbf{What if} there is no sufficient local data to calibrate a culture-compatible AV? Under such situations, these celebrated data-driven approaches will cease to be effective, and the heavy dependence on local data will exacerbate the incompatibility and inequity of AV deployment over time.

\begin{figure}[t!]
    \centering
    \includegraphics[width=1\textwidth]{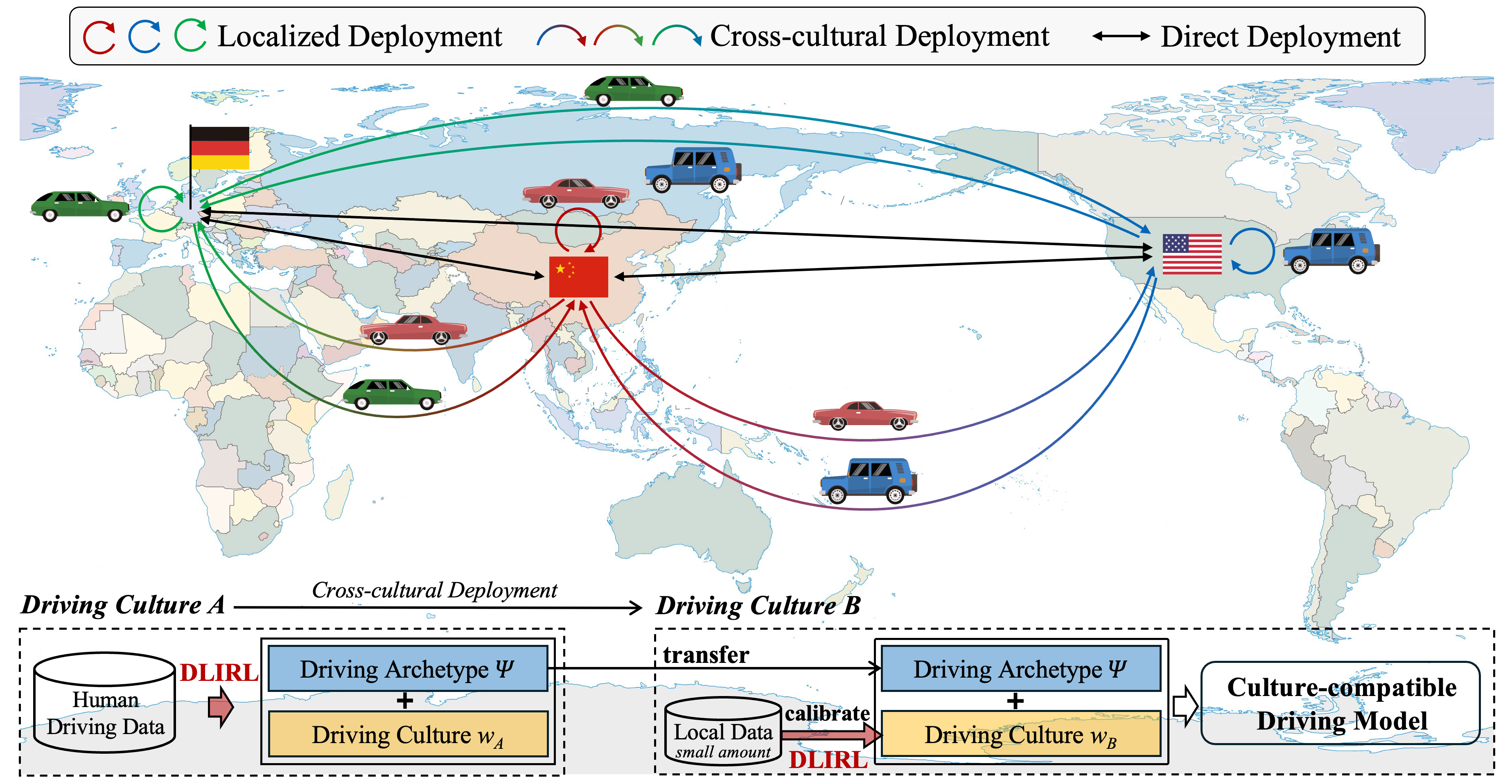}
    \caption{Cross-cultural deployment framework for autonomous vehicles.
The proposed data-light inverse reinforcement learning approach enables cross-cultural autonomous vehicle (AV) deployment. Localized deployment (green) relies heavily on abundant local driving data, while direct deployment (black) attempts to deploy AVs without adapting to local cultures, often leading to incompatibility. Cross-cultural deployment (red and blue) transfers a driving archetype ($\Psi$) extracted from one culture (e.g., Culture A) and calibrates it to another culture (e.g., Culture B) using only a small amount of local data. The driving culture ($w_A$) from Culture A is first decoupled from the archetype ($\Psi$), and the latter is transferred to Culture B. Subsequently, $w_B$ is calibrated based on a small amount of local data in Culture B to produce a culture-compatible driving model. This framework addresses data scarcity challenges and ensures seamless integration of AVs into diverse driving cultures globally. (The source of world map: \url{http://bzdt.ch.mnr.gov.cn/})}
    \label{framework}
\end{figure}

In this study, we propose an expeditious cross-cultural AV deployment scheme named data-light inverse reinforcement learning (DLIRL).
Figure \ref{framework} (bottom) illustrates the process of cross-cultural deployment from Culture A to Culture B. First, based on the driving data in Culture A, DLIRL can decouple human driving behavior into two components: driving archetype $\Psi$ and driving culture $w_A$. Then, we can directly transfer the driving archetype $\Psi$ to Culture B. With DLIRL again and the transferred driving archetype, we can use a small amount of local data to compute the driving culture $w_B$ for Culture B and finally calibrate a culture-compatible driving model.
We validate the effectiveness of our approach on three open-source large-scale naturalistic driving datasets on highways from Germany, China, and the USA, with cumulative testing mileage of over 44,475 km, 6,541 km, and 5,068 km, respectively.
The empirical result demonstrates that our scheme can achieve a decreased data dependence of 98.67\% at best in cross-cultural deployment across the three countries.
That said, only 1.33\% of the local data is now sufficient to calibrate a culture-specific AV compared to what was previously required.
Looking forward, this will bring a broader AV market worldwide, particularly in those regions that lack abundant data to develop a culture-compatible AV, which will help address the global issues of incompatibility and inequity raised by AVs.
% 
%Furthermore, through the lens of our scheme, we unveil the ‘variable’ (called driving culture) and the ‘invariable’ (called driving archetype) hidden in naturalistic driving behaviors, furnishing a plausible understanding of why human drivers who have mastered driving skills within a specific driving culture can swiftly adapt and integrate into other driving cultures.
% 
%This result provides a mechanically plausible explanation for why humans innately excel at few-shot learning (a kind of capability that leverages minimal prior knowledge to generalize to new tasks rapidly \cite{wang2020generalizing}) in the real world.

\section*{Result}\label{sec2}

In this section, we evaluate the performance of DLIRL in cross-cultural AV deployment. As shown in Figure \ref{framework}, three countries, Germany, China, and the USA, along with their culture-specific naturalistic driving datasets, HighD \cite{krajewski2018highd}, Dji \cite{zhang2023ad4che}, and NGSIM \cite{coifman2017critical},  are selected for validation. In this paper, we initially train a driving model using full data from Germany and then test the performance of cross-cultural deployment based on 1.33\% of China data (abbreviated as “\textbf{Germany-to-China}”) and 2.2\% of the USA data (abbreviated as “\textbf{Germany-to-the-USA}”). The other two groups of cross-cultural deployment results can be seen in the Appendix (from China to Germany and the USA, as well as from the USA to Germany and China).

As summarized in Table \ref{tab:deployment}, we consider three deployment schemes and compare them with the original data to evaluate the proposed DLIRL: 
\begin{itemize}
    \item \textbf{Original data} refers to naturalistic driving data from three countries of cultures: DJI for China, HighD for Germany, and NGSIM for the USA. These datasets serve as both training data and validation ground truth.

    \item \textbf{Localized deployment} refers to directly using local data from the target culture to train and deploy the model. Here, we deploy a model based on a long-short-term-memory network (LSTM) as a baseline for comparison.
    %refers to training and deploying a model without cross-cultural adaptation or local data for recalibration. This approach utilizes a standard single-layer LSTM as the baseline and serves as a benchmark for comparison with DLIRL-based methods.
    
    \item \textbf{Direct deployment} involves applying a trained culture-specific model directly to a different culture, without recalibration using local data. Here, we train the model in the original culture using DLIRL and test it directly in other cultural contexts.
    %refers to applying a culture-specific model directly in a different cultural context without using local data for recalibration. This method involves training a model using DLIRL on one culture and testing it in another without recalibration. This approach is designed as a comparison with cross-cultural deployment to test whether a model trained on one culture can effectively adapt to a different culture.
    
    \item \textbf{Cross-culture deployment} refers to transferring a trained culture-specific model to a target culture and recalibrating it using local data from the target culture, thus crafting a culture-compatible driving model. Here, both the model training in the original culture and the recalibration in the target culture are based on the proposed DLIRL.

\end{itemize}

\begin{table}[h]
    \centering
    \begin{tabular}{lcccc}
        \toprule
         & Cross-Culture & Original Data for Training & Local Data for Recalibration & Method \\
        \midrule
        Localized Deployment    & \texttimes & \checkmark & Not Applicable & LSTM \\       
        Direct Deployment       & \checkmark & \checkmark & \texttimes & DLIRL \\
        Cross-Cultural Deployment & \checkmark & \checkmark & \checkmark & DLIRL \\
        \bottomrule
    \end{tabular}
    \caption{Comparison of Different Deployment Approaches}
    \label{tab:deployment}
\end{table}

In the following, we present two sample cases to demonstrate the performance of DLIRL: one for Germany-to-China and the other for Germany-to-the-USA. Next, we conduct a series of large-scale statistical analyses across the full datasets of the three cultures, encompassing three aspects: collective behavioral distribution, driving styles, and driving safety preferences.

\subsection*{Two Sample Cases}

% 在这一节
\begin{figure}[!h]
    \centering
    \includegraphics[width=1\textwidth]{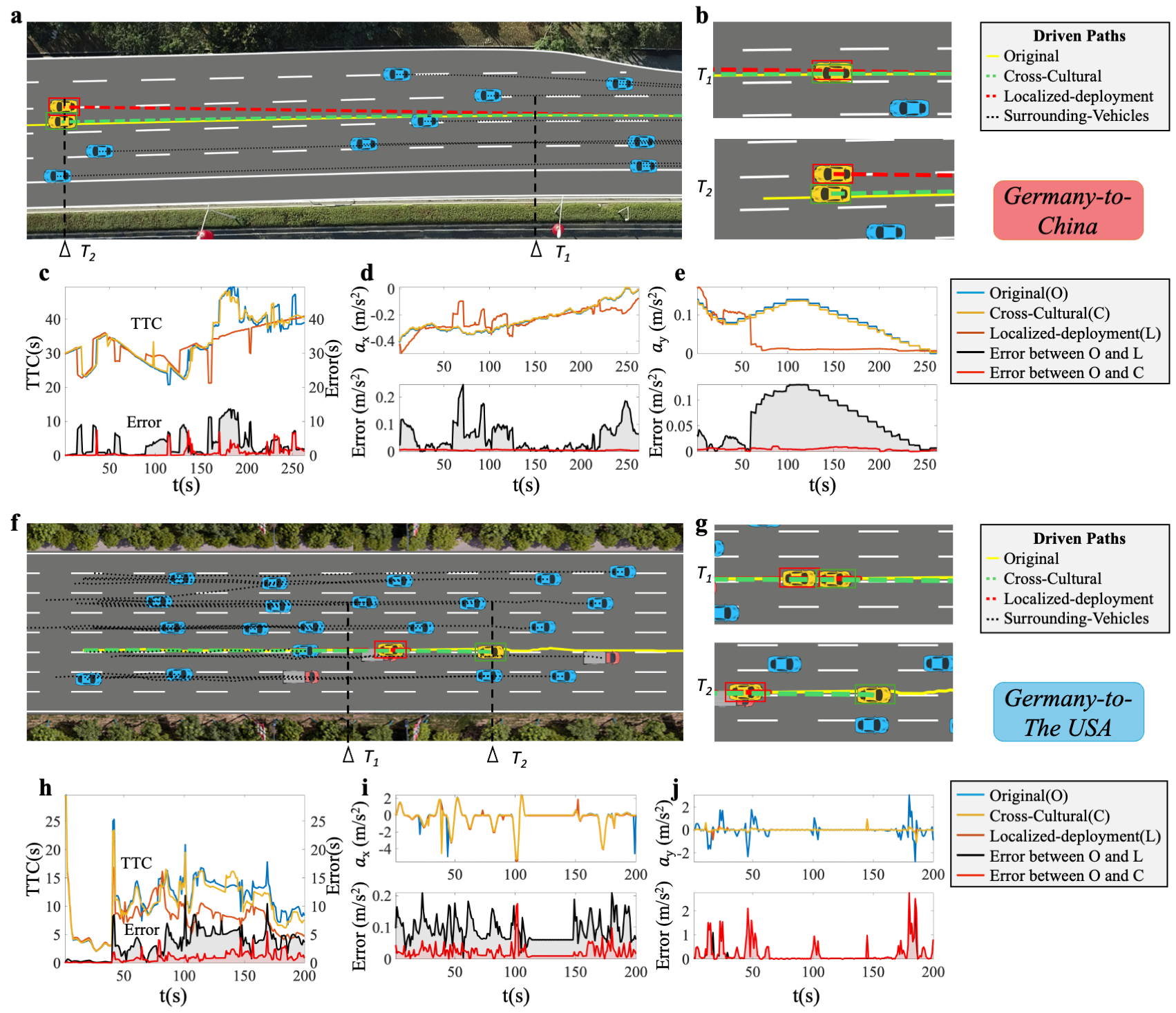}
    \caption{Cross-cultural and localized deployment comparisons for Germany-to-China and Germany-to-the-USA.
\textbf{a}, Scenario for Germany-to-China, where the cross-cultural deployment path closely follows the original data, while the localized deployment deviates.
\textbf{b}, The close-ups at $T_1$ and $T_2$, indicate that both deployments align at $T_1$, but the localized deployment shifts upward at $T_2$.
\textbf{c}, TTC curves show that the cross-cultural deployment maintains closer consistency with the original than the localized deployment.
\textbf{d}, $a_x$ curves indicate that the cross-cultural deployment better preserves the original data, while the localized deployment deviates.
\textbf{e}, $a_y$ curves further confirm that the cross-cultural deployment retains a driving behavior closer to the original.
\textbf{f}, Scenario for Germany-to-the-USA, illustrating that the cross-cultural deployment remains well aligned with the original data, while the localized deployment lags.
\textbf{g}, The close-ups at $T_1$ and $T_2$, show that the localized deployment exhibits noticeable lag at $T_2$.
\textbf{h}, TTC curves indicate that the cross-cultural deployment stays closer to the original data, while the localized deployment exhibits larger deviations.
\textbf{i}, $a_x$ curves show that the cross-cultural deployment more accurately follows the original data, whereas the localized deployment drifts.
\textbf{j}, $a_y$ curves exhibit minimal differences among the deployments, indicating similar lateral driving behavior in this scenario.}
    \label{case}
\end{figure}

In this section, we present the results of cross-cultural deployment using DLIRL: one for Germany-to-China and the other for Germany-to-the-USA. We compare the cross-cultural deployment and the localized deployment, demonstrating their driving paths, close-ups at two specific moments, and the curves of TTC and acceleration. For cross-cultural deployment, we first train the model using the full German data. Then, for the Germany-to-China deployment, we recalibrate the trained model using 1.33\% of the local data from China, while for the Germany-to-the-USA deployment, we use 2.2\% of local data from the USA. In contrast, for localized deployments in both Germany-to-China and Germany-to-the-USA, we train the model directly using 30\% of the local data from each country.

Figures \ref{case}a-e show the results of Germany-to-China, and Figures \ref{case}f-j show the results of Germany-to-the-USA. Figures \ref{case}a and \ref{case}f display the two sample cases. The ego vehicle (EV) in control is marked in yellow. Green dashed lines show the driving path for cross-cultural deployment, red dashed lines are the driving path for localized deployment, and yellow lines are the original path. Surrounding vehicles are blue, with their driving paths shown as black dotted lines. Moments $T_1$ and $T_2$ in Figures \ref{case}a and \ref{case}f are our selected points, and their corresponding scenarios are detailed in Figures \ref{case}b and \ref{case}g. Figures \ref{case}c and \ref{case}h show the TTC curves for the EV. Figures \ref{case}d and \ref{case}i show the $x$-direction acceleration $(a_x)$ curves, and Figures \ref{case}e and j show the $y$-direction acceleration $(a_y)$. Figures \ref{case}c-e and \ref{case}h-j present the statistical results of cross-cultural and localized deployments in comparison with the original data. The cross-cultural deployment results are shown in yellow lines, while the localized deployment results are shown in orange lines. The blue lines represent the original data, the black lines illustrate the differences between the original data and the localized deployment, and the red lines indicate the differences between the original data and the cross-cultural deployment.

First, we analyze the results of Germany-to-China. In Figure \ref{case}a, the cross-cultural deployment path closely aligns with the original data, but the localized deployment path deviates significantly. Figure \ref{case}b shows that at the moment $T_1$, the results of both deployments are similar, but at moment $T_2$, the localized deployed causes the EV to shift upward by 1.9 m in the $y$-direction. In Figure \ref{case}c, the TTC values of cross-cultural match the original data more closely, with a mean error of 0.9477 s, while the localized deployment deviates more significantly, with a mean error of 3.4276 s. Figures \ref{case}d and \ref{case}e show that the acceleration results of cross-cultural are closer to the original data, with mean errors of 0.0054 \( \text{m}/\text{s}^2 \) in $a_x$ and 0.0117 \( \text{m}/\text{s}^2 \) in $a_y$. In contrast, the localized deployment exhibits larger errors, with mean errors of 0.0532 \( \text{m}/\text{s}^2 \) in $a_x$ and 0.0698 \( \text{m}/\text{s}^2 \) in $a_y$. These results show that our method captures driving behavior accurately in the cross-cultural deployment from Germany to China.

Next, we analyze the deployment results of Germany-to-the-USA. Figures \ref{case}f and \ref{case}g show that the cross-cultural deployment path closely aligns with the original data, while localized deployment causes a noticeable lag of 16.2 m. In Figure \ref{case}h, the TTC values of cross-cultural deployment match the original data closely, with a mean error of 0.9228 s, whereas the localized deployment deviates more significantly, with a mean error of 3.3967 s. Figure \ref{case}i shows that the $a_x$ in cross-cultural deployment aligns more closely to the original data, with a mean error of 0.0054 \( \text{m}/\text{s}^2 \), compared to 0.0532 \( \text{m}/\text{s}^2 \) for localized deployment. Figure \ref{case}j shows that the cross-cultural deployment achieves a smaller error in $a_y$, with a mean error of 0.0117 \( \text{m}/\text{s}^2 \), compared to 0.0198 \( \text{m}/\text{s}^2 \) for the localized deployment. These results further confirm the effectiveness of our method in the cross-cultural deployment from Germany to the USA.

In summary, compared to the localized deployment, the proposed cross-cultural deployment achieves more culture-compatible deployment in both cases. In the following, we will present the results of a series of large-scale statistical analyses across three cultures.

\subsection*{Collective Behavioral Distribution}
\begin{figure}[!h]
    \centering
    \includegraphics[width=1\textwidth]{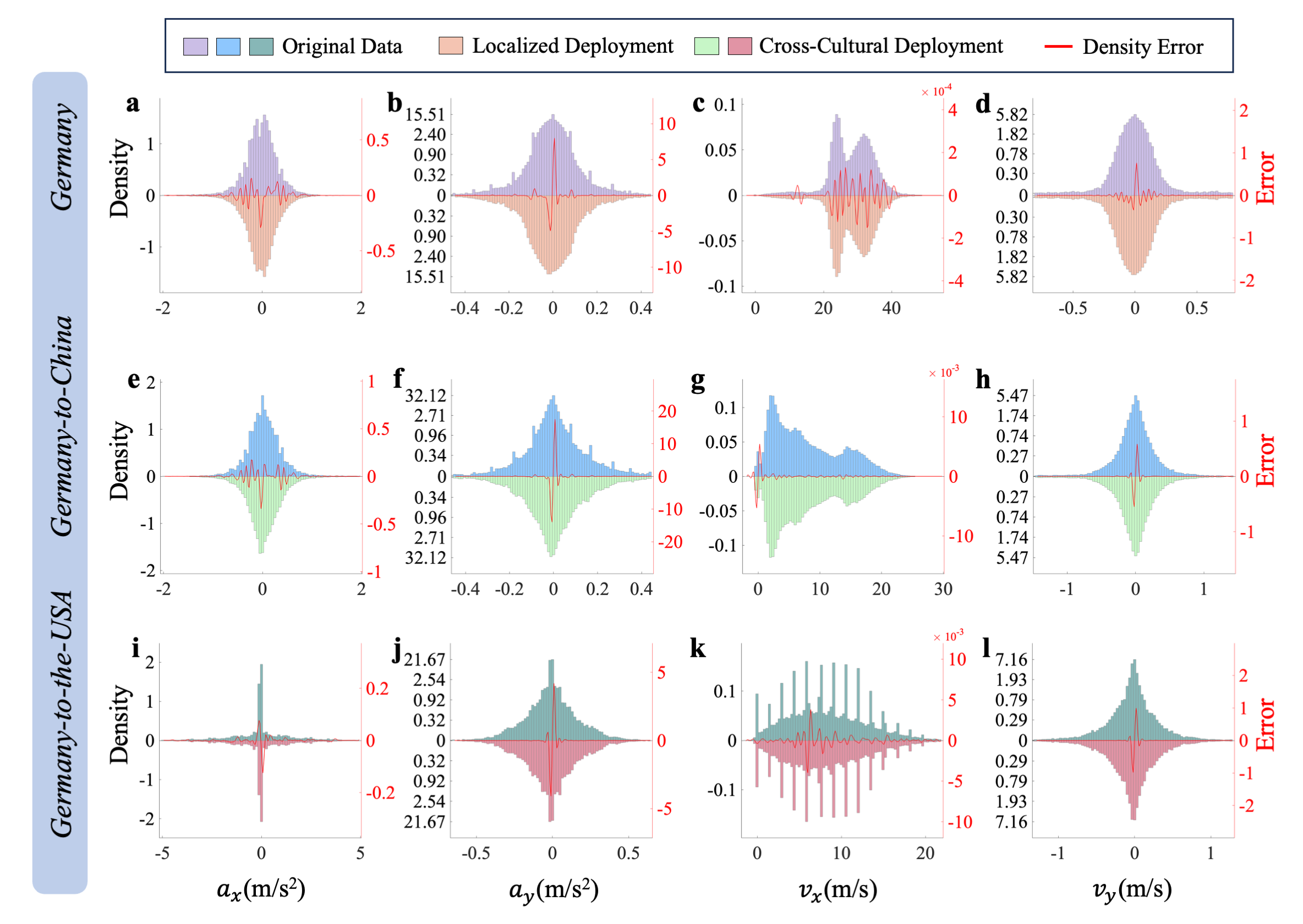} 
    \caption{Comparison of localized and cross-cultural deployment results in driving behavior distributions.
        \textbf{a}–\textbf{d}, Density distributions of $a_x$, $a_y$, $v_x$, and $v_y$ for the localized deployment trained on 100\% German data. The red lines indicate density errors between the localized deployment and the original data, showing strong alignment.
        \textbf{e}–\textbf{h}, Cross-cultural deployment results for Germany-to-China. The distributions demonstrate adaptation to China's driving behavior, with distribution deviations shown in red lines.
\textbf{i}–\textbf{l}, Cross-cultural deployment results for Germany-to-the-USA. The density distributions indicate the model's adaptation to the USA's driving behavior, with distribution deviations shown in red lines.
    }
    \label{fig1}
\end{figure}

\begin{figure}[!h]
    \centering
    \includegraphics[width=0.6\textwidth]{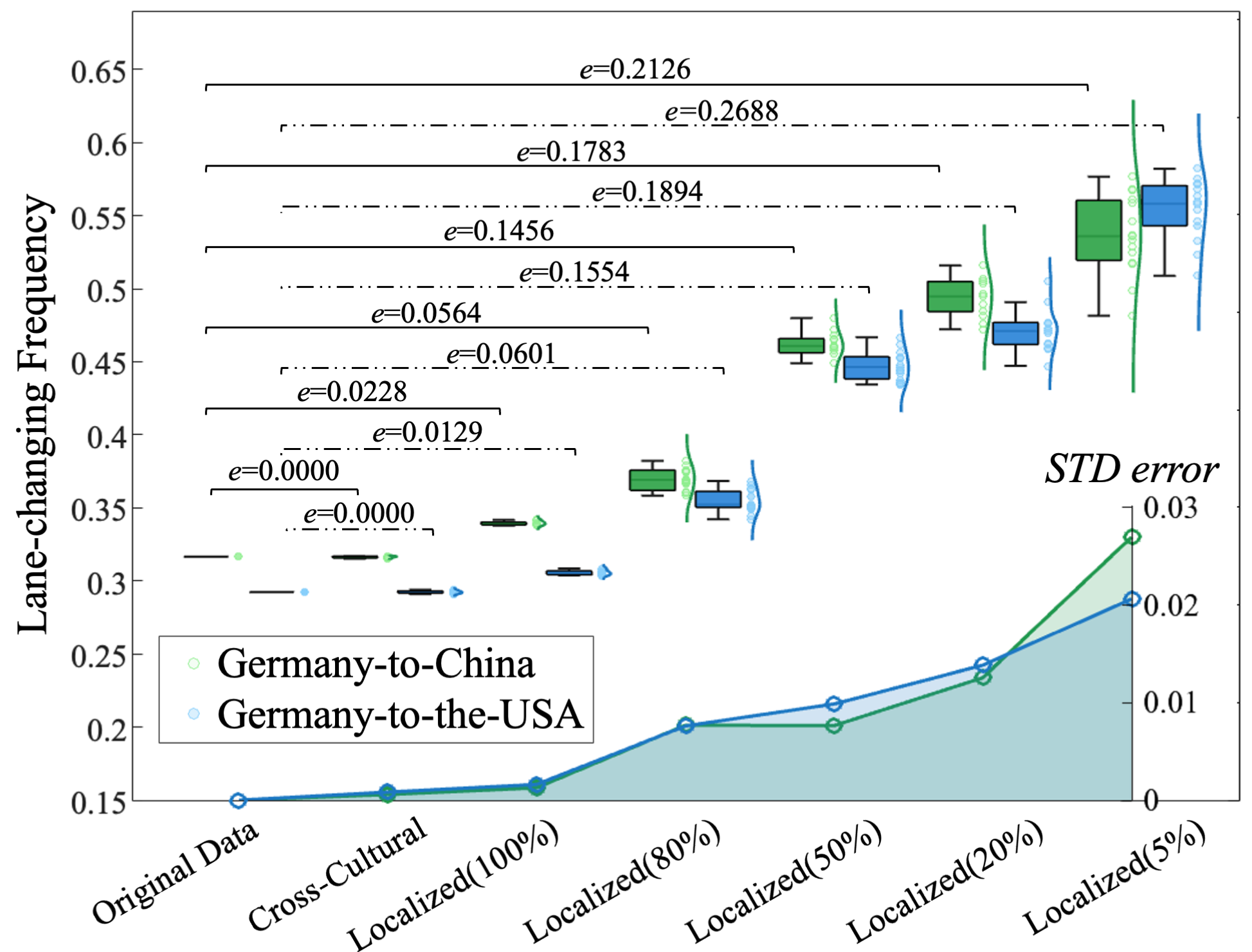}
    \caption{Lane-changing frequency across localized and cross-cultural deployments.
Box plots show the lane-changing frequency distributions for Germany-to-China (green) and Germany-to-the-USA (blue). The cross-cultural deployment closely aligns with the original data, while localized deployment exhibits increasing deviations as training data decreases. The shaded region at the bottom represents the standard error (STD), highlighting the growing instability in lane-changing frequency with reduced local data.}
    \label{fig3}
\end{figure}
%在这一节，我们描述了驾驶行为的分布，具体展示为从速度加速度的总体数学分布，和换道频率分布----overall换一个词
In this section, we present the driving behavior distributions of three cultures, encompassing the statistical profiles of velocity, acceleration, and lane-changing frequency. We evaluate DLIRL’s performance across different driving cultures by comparing the collective driving behavioral distribution under localized and cross-cultural deployment. First, we train the model using data from German culture. Then, we evaluate its performance in localized deployment within Germany and cross-cultural deployment in Germany-to-China and Germany-to-the-USA. As for collective behavior distribution, we primarily focus on $a_x$, $a_y$, $v_x$, and $v_y$. We compare the deviations between localized deployment, cross-cultural deployment, and culture-specific original data. Figure \ref{fig1} illustrates the collective behavioral distribution results for three scenarios: Germany (\ref{fig1}a-d), Germany-to-China (\ref{fig1}e-h), and Germany-to-the-USA (\ref{fig1}i-l). Each group of subfigures corresponds to a specific variable: $a_x$ (\ref{fig1}a, \ref{fig1}e, \ref{fig1}i), $a_y$ (\ref{fig1}b, \ref{fig1}f, \ref{fig1}j), $v_x$ (\ref{fig1}c, \ref{fig1}g, \ref{fig1}k), and $v_y$ (\ref{fig1}d, \ref{fig1}h, \ref{fig1}l). These figures depict the behavioral distribution density of original data and deployment results (indicated on the left y-axis) alongside the deviations between them (indicated on the right y-axis).

% 首先，我们报告三种文化之间的行为分布差异。
We report the difference in behavioral distributions across the three cultures. Figure \ref{fig1}c shows that $v_x$ in Germany primarily falls between 15 and 45 \( \text{m}/\text{s} \). While in cross-cultural deployment for Germany-to-China and Germany-to-the-USA, as shown in Figures \ref{fig1}g and \ref{fig1}k, the velocities are concentrated between 0 and 25 \( \text{m}/\text{s} \). Figure \ref{fig1}d shows that $v_y$ distributions are more consistent across the three cultures, with Germany ranging from -0.3 to 0.3 \( \text{m}/\text{s} \), and Germany-to-China and Germany-to-the-USA ranging from -0.25 to 0.25 \( \text{m}/\text{s} \), as shown in Figures \ref{fig1}h and \ref{fig1}l. For driving acceleration, $a_x$ in Germany exhibits a broader distribution, spanning -0.3 to 0.4 \( \text{m}/\text{s}^2 \), as shown in Figure \ref{fig1}a. The cross-culture deployment results, as shown in Figures \ref{fig1}e and \ref{fig1}i, indicate that Germany-to-China and Germany-to-the-USA exhibit mean accelerations centered around 0.2 \( \text{m}/\text{s}^2 \) and 0.1 \( \text{m}/\text{s}^2 \) respectively. Figure \ref{fig1}b shows that $a_y$ in Germany ranges from -0.15 to 0.15 \( \text{m}/\text{s}^2 \), while in Germany-to-China and Germany-to-the-USA $a_y$ is more constrained, spanning -0.1 to 0.1 \( \text{m}/\text{s}^2 \) and -0.2 to 0.2 \( \text{m}/\text{s}^2 \), respectively, as shown in Figures \ref{fig1}f and \ref{fig1}j. These comparisons highlight significant differences across the three cultures, particularly in longitudinal velocity and acceleration.

Figures \ref{fig1}a-d present the results of the localized deployment under 100\% German data.
% 描述结果
Red lines show the deviation between the localized deployment (100\%) and the original data.
The MSE are 0.0128 \( \text{m}^2/\text{s}^4 \) for $a_x$, 0.0095 \( \text{m}^2/\text{s}^4 \) for $a_y$, 0.0032 \( \text{m}^2/\text{s}^2 \) for $v_x$, and 0.0021 \( \text{m}^2/\text{s}^2 \) for $v_y$, as shown in Figures \ref{fig1}a, \ref{fig1}b, \ref{fig1}c, and \ref{fig1}d, respectively.
% Findings
The results suggest that when local data are fully utilized, the localized method achieves a high-performing alignment with the driving behavior distribution of German culture.
% Cross-culture deployment
Next, we evaluate cross-cultural deployment by comparing the MSE between the cross-cultural deployment results and the culture-specific original data.
% 其他数据集
Specifically, we present the results of Germany-to-China and Germany-to-the-USA in Figures \ref{fig1}e-l.
Figures \ref{fig1}e-h present the deployment result of Germany-to-China, with red lines showing their distribution deviations. A comparison between the original data and the deployment results yields MSE values of 5.5e-7 \( \text{m}^2/\text{s}^4 \) for $a_x$, 5.8e-3 \( \text{m}^2/\text{s}^4 \) for $a_y$, 5.5e-7 \( \text{m}^2/\text{s}^2 \) for $v_x$, and 5.8e-3 \( \text{m}^2/\text{s}^2 \) for $v_y$.
Likewise, Figures \ref{fig1}i-l show the deployment results of Germany-to-the-USA, with red lines marking specific distribution deviations. The MSE results are 4.2e-7 \( \text{m}^2/\text{s}^4 \) for $a_x$ and 1.7e-2 \( \text{m}^2/\text{s}^4 \) for $a_y$. The MSE for $v_x$ is 4.2e-7 \( \text{m}^2/\text{s}^2 \), while for $v_y$, it is 1.7e-2 \( \text{m}^2/\text{s}^2 \).
% 总结
The results indicate that DLIRL effectively facilitates the cross-culture deployment from Germany to China and the USA in terms of acceleration and velocity. Despite the substantial differences in collective driving behavior across cultures, our model accurately captures and aligns with culture-specific characteristics with limited local driving data.

% 这里我们主要是讨论迁移前后的模型在换道频率上的分布
Next, we analyze the lane-changing frequency distributions for localized and cross-cultural deployment.

% 在这一节中，我们展示了在不同比例(包括5%, 20%, 50%，80%，100%)的数据集的模型deployment结果。对每一个Dji和NGSIM的应用结果分别表示为图中的绿色和蓝色。每一个结果（？），基于50次训练于随机抽取对应比例的数据集，其换道频率如散点表示。箱体图展示了statistical results 
We present the results of localized deployment using different dataset proportions (5\%, 20\%, 50\%, 80\%, and 100\%) and cross-cultural deployment to calibrate. Figure \ref{fig3} visualizes the lane-changing frequency results for Germany-to-China (green) and Germany-to-the-USA (blue). Each point represents the outcome of 50 independent training runs, where data were randomly sampled at the corresponding proportions. Scatters beside the box illustrate the frequency distribution, while box plots summarize the statistical results.

% 根据我们的测试结果，迁移模型与原数据的换道频率处于0.32(DJI)与0.28(NGSIM)水平，而使用100%数据集训练的无迁移模型的换道频率为0.31(DJI)和0.29（NGSIM），与原始数据保持了约2%的换道频率误差。
Figure \ref{fig3} illustrates the average lane-changing frequencies for cross-cultural deployment and localized deployment. The cross-cultural deployment results show frequencies of 0.32 for Germany-to-China and 0.29 for Germany-to-the-USA, closely aligning with the original data. In comparison, the localized deployment (100\%) yields slightly lower frequencies of 0.33 for Germany-to-China and 0.30 for Germany-to-the-USA, reflecting a minimal deviation of 0.01 from both the original data and cross-cultural deployment results. This result highlights the effectiveness of our cross-cultural deployment in preserving consistent lane-changing behavior across driving cultures.
% 使用80%数据训练的换道频率为xx(dji)和xx(NGSIM),使用50%数据训练的换道频率为xx(dji)和xx(NGSIM)，使用20%数据训练的换道频率为xx(dji)和xx(NGSIM)，使用5%数据训练的换道频率为xx(dji)和xx(NGSIM)。
For localized deployment in China, models trained with 80\%, 50\%, 20\%, and 5\% local data yield the average lane-changing frequencies of 0.37, 0.46, 0.49, and 0.53, respectively; for the USA, the frequency results are 0.35, 0.44, 0.48, and 0.56.
% 图中下方的曲线展示了每一组数据的均方根误差。随着训练数据的减少，换道频率的结果变得越加不稳定。
The curves at the bottom of Figure \ref{fig3} demonstrate the standard error (STD) for localized deployment. As the local data decreases, lane-changing frequency in localized deployment becomes increasingly unstable and deviates further from the original data.
% % 使用的训练数据集越少则换道频率差别越大。由此可见，我们的迁移方法能够大幅度减少训练数据集依赖，同时能够准确的捕捉到驾驶规则遵守的特征能力。
The localized deployment results show that the less training data is used, the more significant the difference in lane-change frequency.
% 值得注意的是 localized(100%)并没有完全契合原始数据而我们的方法打成了这一结果, 这是由于我们的方法 xxxx
Notably, the localized deployment (100\%) fails to fully align with the original data because traditional deep learning models weaken the influence of early input as new data arrives, making it difficult to maintain accurate mapping across all data points and leading to underfitting. In contrast, our method achieves this result by transferring and leveraging the archetype for rapid recalibration during cross-cultural deployment. This enables our approach to fit the data well even with a small amount of local data.

Overall, compared to localized deployment, the proposed DLIRL method effectively captures lane-changing patterns while significantly reducing the dependence on local data. By leveraging cross-cultural deployment, our method maintains high accuracy across different driving cultures, demonstrating its robustness and efficiency in cultures without sufficient local data.

\subsection*{Driving Styles}

\begin{figure}[h]
    \centering
    \includegraphics[width=1\textwidth]{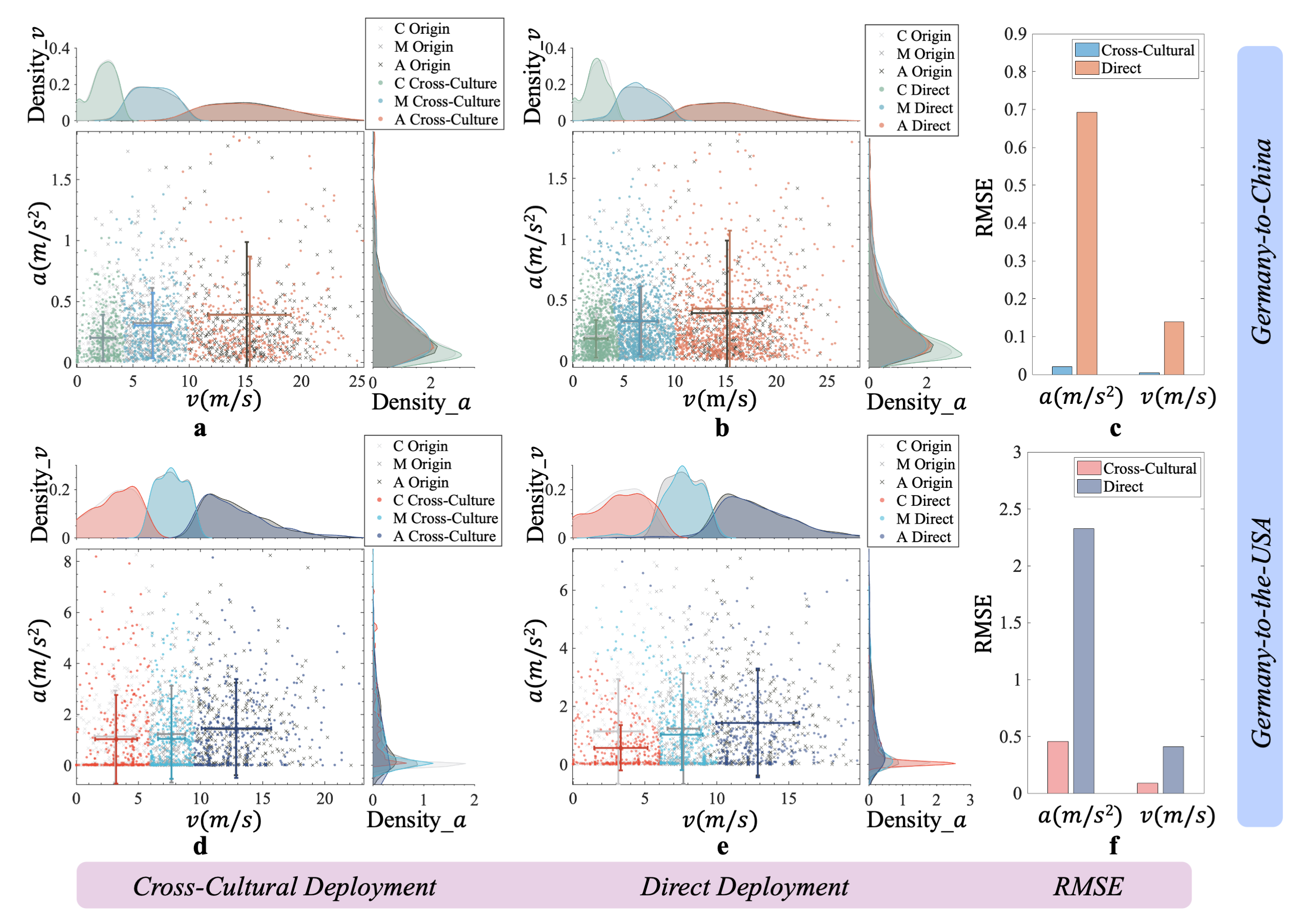} 
    \caption{Comparison of cross-cultural and direct deployment in capturing driving styles.
\textbf{a}, \textbf{b}, Velocity-acceleration distributions for Germany-to-China under cross-cultural deployment (\textbf{a}) and direct deployment (\textbf{b}). Scatter points represent individual driving styles, with density distributions shown along the axes.
\textbf{c}, RMSE of velocity and acceleration for Germany-to-China, showing lower errors in cross-cultural deployment.
\textbf{d}, \textbf{e}, Velocity-acceleration distributions for Germany-to-the-USA under cross-cultural deployment (\textbf{d}) and direct deployment (\textbf{e)}.
\textbf{f}, RMSE for Germany-to-the-USA, indicating that cross-cultural deployment better preserves driving styles.
}
    \label{fig2}
\end{figure}

In this section, we evaluate the performance of DLIRL in capturing individual driving styles for Germany-to-China and Germany-to-the-USA. Driving styles are categorized into three types based on velocity and acceleration: aggressive, moderate, and conservative \cite{lyu2022using}. Specifically, we calculate the squared sums of each individual’s average velocity and acceleration, sort the results, and uniformly split them into three groups corresponding to the three driving styles \cite{liu2019research}. 
Figure \ref{fig2} presents the overall deployment results in driving styles, encompassing C(conservative), M(moderate), and A(aggressive), with Figures \ref{fig2}a, \ref{fig2}b, and \ref{fig2}c showing results for Germany-to-China, and Figures \ref{fig2}d, \ref{fig2}e, and \ref{fig2}f showing results for Germany-to-the-USA. In Figures \ref{fig2}a,  \ref{fig2}b,  \ref{fig2}d, and \ref{fig2}e, the scatter plots illustrate individual driving behaviors, with three driving styles represented by three colors. 
Specifically, in Figures \ref{fig2}a and \ref{fig2}b, the conservative, moderate, and aggressive driving styles are shown in green, blue, and red, respectively. In Figures \ref{fig2}d and \ref{fig2}e, the same driving styles are represented: red for conservative, blue for moderate, and dark blue for aggressive. The distributions of velocity and acceleration are shown on the top and right axes in each figure. 
The gray x-shaped scatter points represent the three driving styles of the original data. Circular scatter points in Figures \ref{fig2}a and \ref{fig2}d show the results of cross-cultural deployment, while Figures \ref{fig2}b and \ref{fig2}e show the result of direct deployment. The average values for the three driving styles are highlighted with crosshairs.

% 我们首先说明两种部署方法之间个人驾驶风格的差异。
We first evaluate the performance of DLIRL for Germany-to-China. Figures \ref{fig2}a and \ref{fig2}b illustrate the three driving styles under cross-cultural and direct deployment, respectively. Figure \ref{fig2}a shows the results of cross-cultural deployment. The grey scatters represent the original data, where the average values of velocity and acceleration for conservative, moderate, and aggressive styles are 2.3 \( \text{m}/\text{s} \) and 0.20 \( \text{m}/\text{s}^2 \), 6.7 \( \text{m}/\text{s} \) and 0.32 \( \text{m}/\text{s}^2 \), and 15.1 \( \text{m}/\text{s} \) and 0.39 \( \text{m}/\text{s}^2 \), respectively. Under cross-cultural deployment, the corresponding average values are 2.2 \( \text{m}/\text{s} \) and 0.2 \( \text{m}/\text{s}^2 \), 6.6 \( \text{m}/\text{s} \) and 0.3 \( \text{m}/\text{s}^2 \), and 15.4 \( \text{m}/\text{s} \) and 0.39 \( \text{m}/\text{s}^2 \). The velocity discrepancies between the original data and cross-cultural deployment are 0.1 \( \text{m}/\text{s} \), 0.1 \( \text{m}/\text{s} \), and 0.3 \( \text{m}/\text{s} \) for conservative, moderate, and aggressive styles, respectively, while the acceleration discrepancies are 0.0 \( \text{m}/\text{s}^2 \), 0.02 \( \text{m}/\text{s}^2 \), and 0.0 \( \text{m}/\text{s}^2 \). Figure \ref{fig2}b illustrates the results of direct deployment. Grey scatters again represent the original data, where the average values for conservative, moderate, and aggressive styles remain at 2.3 \( \text{m}/\text{s} \) and 0.20 \( \text{m}/\text{s}^2 \), 6.7 \( \text{m}/\text{s} \) and 0.32 \( \text{m}/\text{s}^2 \), and 15.1 \( \text{m}/\text{s} \) and 0.39 \( \text{m}/\text{s}^2 \), respectively. 
Under direct deployment, the corresponding average values shift to 2.2 \( \text{m}/\text{s} \) and 0.18 \( \text{m}/\text{s}^2 \), 6.5 \( \text{m}/\text{s} \) and 0.32 \( \text{m}/\text{s}^2 \), and 15.4 \( \text{m}/\text{s} \) and 0.43 \( \text{m}/\text{s}^2 \). Compared to the original data, the velocity discrepancies are 0.1 \( \text{m}/\text{s} \), 0.2 \( \text{m}/\text{s} \), and 0.3 \( \text{m}/\text{s} \) for conservative, moderate, and aggressive styles, respectively, while the acceleration discrepancies are 0.02 \( \text{m}/\text{s}^2 \), 0.0 \( \text{m}/\text{s}^2 \), and 0.04 \( \text{m}/\text{s}^2 \). 
% 可能还要加文字，因为虽然平均值更接近但是区间没有crosscultural接近
These results indicate that direct deployment results in greater deviations in driving styles, especially for aggressive driving. Specifically, the velocity and acceleration discrepancies are 0.3 \( \text{m}/\text{s} \) and 0.04 \( \text{m}/\text{s}^2 \) for direct deployment, compared to 0.3 \( \text{m}/\text{s} \) and 0.0 \( \text{m}/\text{s}^2 \) for cross-cultural deployment. This result highlights the effectiveness of DLIRL in capturing driving styles. To better qualify the distributional discrepancy, we calculate the RMSE of velocity and acceleration separately. As shown in Figure \ref{fig2}c, the RMSE values for velocity and acceleration under cross-cultural deployment are 0.004 \( \text{m}/\text{s} \) and 0.02 \( \text{m}/\text{s}^2 \), while under direct deployment, the RMSE values are 0.13 \( \text{m}/\text{s} \) and 0.7 \( \text{m}/\text{s}^2 \). These results demonstrate that our method can effectively capture driving styles compared to direct deployment, which exhibits significant deviations.

Next, we evaluate the performance of cross-cultural deployment for Germany-to-the-USA. Figures \ref{fig2}d and \ref{fig2}e illustrate the three driving styles under cross-cultural and direct deployment, respectively.
Figure \ref{fig2}d shows the results of cross-cultural deployment. The grey scatter points represent the original data, where the average velocity and acceleration for the conservative, moderate, and aggressive driving styles are 3.1 \( \text{m}/\text{s} \) and 1.1 \( \text{m}/\text{s}^2 \), 7.7 \( \text{m}/\text{s} \) and 1.2 \( \text{m}/\text{s}^2 \), and 12.8 \( \text{m}/\text{s} \) and 1.4 \( \text{m}/\text{s}^2 \), respectively. Under cross-cultural deployment, these values shift to 3.2 \( \text{m}/\text{s} \) and 1.0 \( \text{m}/\text{s}^2 \), 7.7 \( \text{m}/\text{s} \) and 1.0 \( \text{m}/\text{s}^2 \), and 12.8 \( \text{m}/\text{s} \) and 1.5 \( \text{m}/\text{s}^2 \). The velocity discrepancies are 0.1 \( \text{m}/\text{s} \), 0.0 \( \text{m}/\text{s} \), and 0.0 \( \text{m}/\text{s} \), while the acceleration discrepancies are 0.1 \( \text{m}/\text{s}^2 \), 0.2 \( \text{m}/\text{s}^2 \), and 0.1 \( \text{m}/\text{s}^2 \).
Figure \ref{fig2}e shows the results of direct deployment. Direct deployment results in higher average values of 3.3 \( \text{m}/\text{s} \) and 0.6 \( \text{m}/\text{s}^2 \), 7.6 \( \text{m}/\text{s} \) and 1.0 \( \text{m}/\text{s}^2 \), and 12.8 \( \text{m}/\text{s} \) and 1.4 \( \text{m}/\text{s}^2 \) for the three styles. Compared to the original data, the velocity discrepancies increase to 0.2 \( \text{m}/\text{s} \), 0.1 \( \text{m}/\text{s} \), and 0.0 \( \text{m}/\text{s} \), while the acceleration discrepancies reach 0.5 \( \text{m}/\text{s}^2 \), 0.2 \( \text{m}/\text{s}^2 \), and 0.0 \( \text{m}/\text{s}^2 \). Overall, direct deployment leads to larger deviations, particularly for conservative driving, where the velocity and acceleration discrepancies reach up to 0.1 \( \text{m}/\text{s} \) and 0.5 \( \text{m}/\text{s}^2 \), compared to 0.1 \( \text{m}/\text{s} \) and 0.1 \( \text{m}/\text{s}^2 \) under cross-cultural deployment. 
As shown in Figure\ref{fig2}f, the RMSE values for cross-cultural deployment are 0.09 \( \text{m}/\text{s} \) for velocity and 0.46 \( \text{m}/\text{s}^2 \) for acceleration, whereas direct deployment results in significantly higher RMSE values of 2.3 \( \text{m}/\text{s} \) for velocity and 0.4 \( \text{m}/\text{s}^2 \) for acceleration. These results also confirm that cross-cultural deployment better captures the target driving styles.

In summary, the proposed cross-cultural deployment method demonstrates superior performance in capturing driving styles compared to direct deployment. It significantly reduces reliance on local data while achieving a closer alignment with the target culture.

\subsection*{Driving Safety Preference}
% 这里我们研究了不同文化下驾驶员对于驾驶安全的倾向和保守程度

% 在这一节
\begin{figure}[h]
    \centering
    \includegraphics[width=0.85\textwidth]{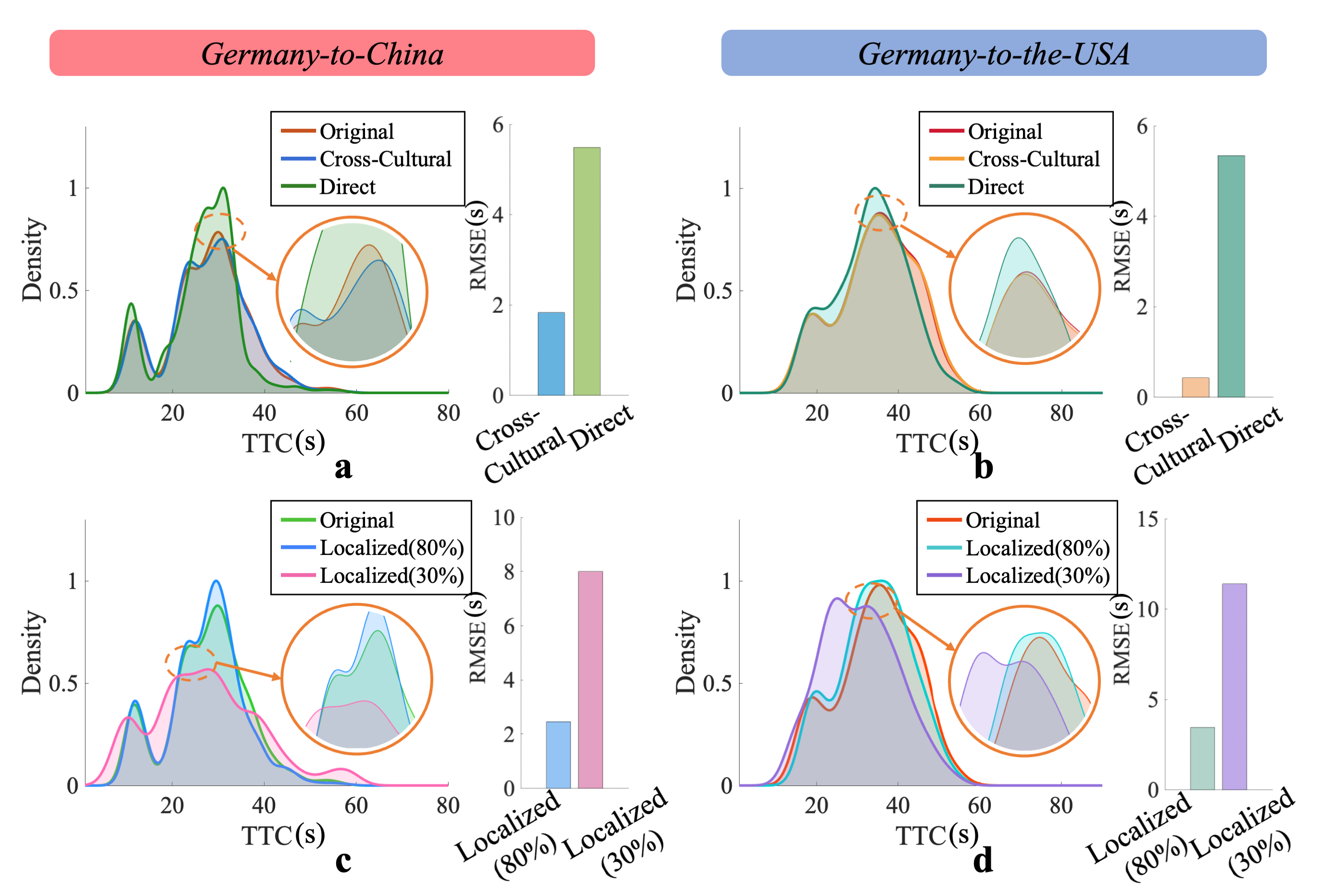}
    \caption{Comparison of TTC distributions in cross-cultural and localized deployments.
\textbf{a}, \textbf{b}, TTC density distributions for Germany-to-China (\textbf{a}) and Germany-to-the-USA (\textbf{b}) under cross-cultural and direct deployment. Cross-cultural deployment closely aligns with the original data, while direct deployment exhibits larger discrepancies. The right-side bar plots show RMSE values, indicating that cross-cultural deployment achieves lower error.
\textbf{c}, \textbf{d}, TTC density distributions for Germany-to-China (\textbf{c}) and Germany-to-the-USA (\textbf{d}) under localized deployment with 80\% and 30\% training data. Reducing local data leads to increasing deviations, with RMSE values reflecting higher errors in low-data scenarios.
}
    \label{fig4}
\end{figure}

% 本节我们用TTC作为指标，评估了迁移模型与HighD模型下驾驶员对于驾驶安全的倾向和保。
In this section, we evaluate driving safety preferences under cross-cultural and localized deployments. We use TTC (Time to Collision) as a measurement and RMSE to analyze the results quantitatively.
% 我们从china,usa数据集中分别选取了100个轨迹，计算其跨文化部署与直接部署的结果，并且将每帧驾驶员对周围环境的反馈以ttc密度的结果呈现在下图中。右图的rmse值反应了跨文化部署、直接部署与原数据的定量误差。
We randomly select 100 cases from China and the USA to compare the results of cross-cultural, direct, and localized deployments. Figure \ref{fig4} shows the deployment results and RMSE values. We present the driver’s safety preferences to the surrounding vehicles based on TTC per frame and present the results as a density curve.

First, we evaluate the performance of DLIRL by analyzing TTC distributions for Germany-to-China. As shown in Figure \ref{fig4}a, the original TTC density (orange line) peaks at 11.9 s and 29.6 s. The cross-cultural deployment results (blue line) align closely with the original distribution, with peaks at 12.1 s and 30.2 s. In contrast, the direct deployment (green line) exhibits noticeable discrepancies, overestimating the density near 10.0 s and misaligning at 30.0 s. The RMSE values further quantify these differences: the cross-cultural deployment yields an RMSE of 1.8 s, while direct deployment results in a significantly higher RMSE of 5.5 s.
Similarly, we evaluate the results for Germany-to-the-USA, as shown in Figure \ref{fig4}b. The original TTC density (pink line) peaks at 17 s and 35 s. Cross-cultural deployment (yellow line) closely follows the original data, maintaining similar peaks with slight underestimations. However, the direct deployment (green line) shows substantial deviations, particularly overestimating densities at both peaks with 0.38 and 0.86. The RMSE for cross-cultural deployment is 0.42 s, while direct deployment yields a larger RMSE of 5.3 s.
These results demonstrate that cross-cultural deployment better captures driving safety preferences compared to culture-specific direct deployment.

We further compare the results of localized deployment using different training data proportions (80\% and 30\%). For Germany-to-China, Figure \ref{fig4}c shows that the localized deployment (80\%) on the blue line closely aligns with the original distribution but exhibits a slightly lower peak. However, the localized deployment (30\%) on the pink line underestimates the main peak, reaching a value of only 0.54 at its highest point. The RMSE values in the bar chart are 2.4 s for localized deployment (80\%) and 8.0 s for localized deployment (30\%).
For Germany-to-the-USA, as illustrated in Figure \ref{fig4}d, the localized deployment (80\%) on the aqua line maintains similar peaks but with reduced prominence, while the localized deployment (30\%) on the purple line exhibits a larger deviation in peak position, which shifts to 24.5 s. The RMSE values highlight a significant difference, with the localized deployment (80\%) achieving an RMSE of 3.4 s compared to 11.4s for the localized deployment (30\%).
Notably, the RMSE of localized deployment with a larger proportion of local data is close to that of cross-cultural deployment. These results indicate that our approach maintains accuracy in capturing driving safety preferences while significantly reducing the reliance on local data.

Overall, higher alignments in cross-cultural deployment compared to direct and localized deployment highlight the robustness of cross-cultural deployment in maintaining driving safety preferences while reducing the reliance on local data.

\section*{Discussion}\label{sec3}
%% 数据驱动
% 先简要总结一下
This study initially investigates the deployment of AVs across diverse driving cultures and proposes an effective data-driven approach to achieve it. The experimental results indicate a significant reduction in data dependency when calibrating a culture-compatible AV, which holds great promise for facilitating the broader deployment of AVs on a global scale, particularly in regions without abundant local data for calibration. This is expected to bring a broader international AV market. 
Additionally, in line with what has been highlighted in \cite{bhoopchand2023learning} to develop a few-shot cultural transmission, our study exemplifies an effective data-driven practice that can well pinpoint and capture the implicit pattern hidden behind driving behaviors, showcasing a promising pathway to investigate and deliberate behavior culture beyond hard coding. 
From a broader context of machine culture, our study examines the social and cultural attributes that AVs, as intelligent machines, should consider and integrate, rather than serving as cold entities in the future \cite{brinkmann2023machine}. 

%% 驾驶文化
At the outset of this paper, we introduced our definition of driving culture. It is worth noting, however, that driving culture is a complex and multifaceted concept lacking universally acknowledged or all-encompassing frameworks \cite{sagberg2015review}. 
It is not our intention to perfectly define driving culture in this paper, so we have approached this concept from a collective behavioral perspective, instead of prescribing its specific contents and details.
Here we focus on a collective behavioral pattern followed by most drivers, differing from the driving habit widely discussed in the scope of individuals \cite{li2021extraction}.
Also, some possibly related physical factors fall outside the scope of this study, such as traffic regulations, traffic signs, and road infrastructure, which are also considered components of driving culture \cite{nordfjaern2014culture}.
In experimental tests, we consider and investigate several subordinate facets of driving culture, including collective behavioral distribution, driving style, adherence to traffic regulation, and driving safety; all of them are commonplace in human driving yet cannot be fully addressed by man-made rules.
Here we call for a well-orchestrated, holistic definition or framework for driving culture, such that future lines can further discuss and formalize which facets should be implicated in driving culture.

%% 其次，我们谈及到了驾驶行为中的变与不变
%
In this study, we decouple driving behavior into two components: driving culture and driving archetype, corresponding to the 'variable' and the 'invariable' in human behavior. 
This explanation stems from a wealth of studies about human behaviors in which human transfer learning \cite{malloy2023accounting} and few-shot learning \cite{song2023comprehensive} are two cruxes widely discussed all the while.
Specifically, the primary objective of transfer learning is to identify the homogeneous and generalizable commonality that collective behaviors follow, which can be regarded as transferable knowledge or a shared behavioral pattern among most individuals \cite{zhuang2020comprehensive,bhaskar2019analyzing}.
Next, given something heterogeneous in new tasks, these shared commonalities can be treated as prior knowledge and then come into play in few-shot learning, thus generating task-specific novel strategies.
Contextualizing this picture into driving scenarios, we use driving archetypes to address the question of "what is transferable commonality," and then employ driving culture to answer "how to operationalize few-shot learning across different driving contexts." 
In light of this, the expeditious adaptation to diverse driving cultures shown in this paper can exemplify the remarkable adaptive capabilities of humans in various contexts.

Our method also enjoys strong scalability. The current model outputs acceleration, which is directly used for AV control. In the future, our approach can function as a module, working alongside more advanced planners or controllers to enable reliable and culture-sensitive driving behavior. 
In addition, there are some spaces for further improvement. 
First, more than the three cultures and highway scenarios, more cultures and scenarios can be considered to validate our approach in future work. Second, exploring how driving culture and traffic regulations collectively shape driving behavior could provide valuable insights into human driving and further support culture-compatible AV deployment. Third, our approach may fall short in regions without local data, such as remote rural areas or countries with underdeveloped AV technology. Addressing this issue will contribute to a more widespread and equitable deployment of autonomous driving in a global context.

\section*{Methods}\label{sec4}

In this study, we develop a cross-cultural deployment scheme for AVs, called DLIRL. Below, we detail how driving culture is taken into account and how to operationalize our method, following from three sections:

\begin{enumerate}
    \item \textbf{Driving Culture}: Driving Culture represents the collective driving behavioral pattern within a specific region, suggesting how local drivers interact with each other and their environments. This section details four key facets that can be absorbed into driving culture, including behavioral distribution, driving style, regulatory compliance, and driving safety.

    \item \textbf{Data-light Inverse Reinforcement Learning}: DLIRL is crafted to decouple driving behavior into two components: driving culture, reflecting region-specific collective behavioral patterns, and driving archetype, capturing the behavioral commonality shared across drivers.
    This section illustrates how DLIRL works in practice based on naturalistic driving data, along with its network architecture. 

    \item \textbf{Cross-cultural Adaptation}: In closing, we illustrate how to capitalize on the extracted driving archetype across driving cultures of different regions. Without extensive retraining, a significantly reduced amount of local data is now sufficient to calibrate a culture-compatible AV in cross-cultural deployment.

\end{enumerate}

\subsection*{Driving Culture}

In this study, four facets fall into consideration when examining driving culture: driving culture: behavioral distribution, driving style, regulatory compliance, and safety awareness. Below is how we measure them across three countries.

\subsubsection*{Behavioral Distribution}
% 是什么
Behavioral distribution refers to the driving locomotion in response to various driving conditions. Acceleration $a$, which directly reflects how drivers manipulate the vehicle, and speed $v$, which serves as an intuitive result of vehicle operation, are both considered in the analysis of behavioral distribution in their lateral (y-axis) and longitudinal (x-axis) dimensions. In doing so we can analyze driving behaviors comprehensively, such as speeding up, slowing down, and steering, thus capturing how drivers interact with their surroundings while driving.

% 换道频率
Furthermore, considering that lane-changing behavior is an important characteristic for measuring high-speed driving behavior, we also take lane-changing frequency $\mathcal{F}_\text{lane-change}$ into account here, calculated by:
\begin{equation}
    \mathcal{F}_\text{lane-change} = \frac{N_c}{N_t}
\end{equation}
where $N_c$ is the number of trajectories that involve lane-changing and $N_t$ is the total number of trajectories in the dataset. This measurement can quantify how often drivers are prone to change lanes, showcasing a collective behavioral feature specific to highways.

\subsubsection*{Driving style}

Driving style is defined as the characteristic manner in which drivers typically operate their vehicles. It is often categorized into three main types: aggressive, moderate, and conservative driving styles, as supported by previous studies. These styles capture the habitual patterns that drivers exhibit over time, encompassing how they manage speed, acceleration, and safety margins. In our research, we perform a clustering analysis on the velocity and acceleration data. Velocity serves as the primary indicator of driving style, representing how drivers typically manage their speed. Acceleration reflects the overall smoothness or abruptness of a driver’s style. We calculate the sum of squares for both velocity and acceleration for each data sample. This allows us to quantify the overall intensity of driving behavior for each sample in terms of speed management and acceleration patterns. Based on these computed values, we manually assign data points to three distinct classes $C_1, C_2, C_3$, corresponding to aggressive, moderate, and conservative driving styles. This can be explained by the following formula:
\begin{equation}
C_k = \left\{ \mathbf{\gamma}i \mid \left( \frac{v_i^2}{\sigma_v^2} + \frac{a_i^2}{\sigma_a^2} \right) \in \left[ \lambda_{k-1}, \lambda_k \right) \right\}, \quad k = 1, 2, 3
\end{equation}
where vector  $\mathbf{\gamma}_i = (v_i, a_i)$  represents the $i$-th sample’s velocity and acceleration, normalized by their standard deviations. The sum of squared components is used to classify samples into clusters  $C_k$  based on thresholds  $\lambda_k$.

This data-driven step helps us validate our manual categorization and uncover additional nuances within and across regions, providing a comprehensive understanding of common driving patterns.

\subsubsection*{Driving Safety}

Driving safety measures how drivers prioritize safety in their driving behavior. Here we use a well-known metric, time-to-collision (TTC), to evaluate driving safety \cite{nadimi2020evaluation}. As a key indicator of how drivers manage risk on the road, TTC measures the remaining time before a potential collision occurs, assuming that the current speed is maintained. Drivers with consistently shorter TTC values tend to exhibit a more risk-tolerant or aggressive style, while those with longer TTC values prioritize maintaining safe distances from other vehicles, reflecting a more cautious approach. By analyzing TTC across different regions, we can capture how safety consciousness varies between cultures, highlighting whether drivers tend to take more calculated risks or emphasize accident avoidance. Understanding this aspect of driving safety allows us to gauge the risk management strategies inherent in different driving cultures, and it also informs the design of autonomous vehicles to adapt accordingly in diverse driving environments where safety expectations may differ. Specifically, we calculate TTC for vehicles in eight different directions relative to the target vehicle: front, back, left, right, and the four diagonal directions. For directions without corresponding surrounding vehicles, we assign a value of -1 to indicate the absence of risk. The total TTC for each direction is computed as the algebraic sum of the individual TTC values along the $x$ and $y$ axes:
\begin{equation}
ttc_\text{total} = \frac{1}{\sum_{i=1}^{8} \mathbb{I}(c)} \sum_{i=1}^{8} \left( \frac{d^x_{i}}{|\Delta v^x_{i}|} + \frac{d^y_{i}}{|\Delta v^y_{i}|} \right) \cdot \mathbb{I}(c)
\end{equation}
% 说明参数
where $d^x_{i}$ and $d^y_{i}$ are the distances between the target vehicle and the surrounding vehicle in the $x$ and $y$ directions respectively, $\Delta v^x_{i}$ and $\Delta v^y_{i}$ are the relative velocities in the $x$ and $y$ directions, and  $\mathbb{I}(c)$ indicates the presence of surrounding vehicles in specified direction.
% 总结
Understanding this aspect of driving safety allows us to gauge the risk management strategies inherent in different driving cultures, and it also informs the design of autonomous vehicles to adapt accordingly in diverse driving environments where safety expectations may differ.

\subsection*{Data-light Inverse Reinforcement Learning} 

In this study, we propose a data-light inverse reinforcement learning framework to reasonably infer the action-value function in no-reward situations, in order to guide human driving behavior. The following sections detail the construction of the no-reward action-value function and the model architecture.

\subsubsection*{Algorithm Illustration}

Human driving can be defined as a Markov Decision Process (MDP) characterized by a reward function $R: S \times A \to \mathbb{R}$. $S$ is the state space and $A$ is the action space. The driver seeks to maximize the expected discounted sum of rewards $G^R$ through a policy $\pi$ that can map $S$ to a probability distribution over $A$, i.e., $\pi: S \to A$. The action-value function under $\pi$ represents the expected cumulative reward with an initial state $s$ and action $a$, which can be expressed as: 
\begin{equation}
    Q^{\pi}(s, a) = \mathbb{E}^{\pi} [G^R | s_0 = s, a_0 = a]
\label{eq4}
\end{equation}

Despite MDP providing a solid theoretical foundation for interpreting driving behavior, the reward function $R$ in an open world is too implicit to formalize clearly. To address this, we borrow from the successor feature framework that can capture driving behavior without explicit reward functions \cite{filos2021psiphi}. Specifically, within the successor feature framework, the one-step reward function can be divided into two components: state-action cumulants $\Phi(s, a)$ and preference vector $w$, by:
\begin{equation}
R^w(s,a) = \Phi(s,a)^{\top} \cdot w
\end{equation}
where $\Phi(s, a)$ captures the one-step state-action relationship and $w$ represents its preference-specific refinement and molding. In doing so we can circumvent the constraints of MDP that rely on explicit rewards to understand driving behavior. Below is how we contextualize this theory in detail.

First, we formulate the driving archetype based on cumulants. Given that driving behavior is continuous, the discounted sum to cumulants $\Psi^\pi(s, a)$ is introduced. Specifically, starting from the initial state $s_0 = s$ and initial action $a_0 = a$ under policy $\pi$, the cumulants can be calculated by:
\begin{equation}
\Psi^\pi(s, a) = \mathbb{E}_{\pi} \left[ \sum_{t=0}^{\infty} \gamma^t \Phi(s_t, a_t) \mid s_0 = s, a_0 = a \right]
\end{equation}
where $\gamma$ is the discount factor (0.9 in this paper). Given a driving dataset, $\Psi^\pi(s, a)$ can extract the collective, shared state-action relationship followed by human drivers as the driving archetype. In our approach, $\Psi^\pi(s, a)$ is represented by a neural network, with an output vector sized by 12×1 (see Section \ref{sec422}).

Then, to capture the influence of specific driving culture on driving behavior, $w$ is used to represent specific driving culture. Specifically, as with $\Psi^\pi(s, a)$, $w$ is also a 12×1 vector, which is learned and updated under a certain driving archetype. The $w$ update process can be expressed by:
\begin{equation}
w_t = w_{t-1} - \frac{\alpha}{\sqrt{\hat{v}_t} + \epsilon} \hat{m}_t
\end{equation}
where $\alpha$ is the learning rate (0.01 in this paper), $\hat{m}_t$ and $\hat{v}_t$ respectively represent the mean and squared mean of gradients at time step $t$, and $\epsilon$ is a constant to avoid division by zero. This update process ensures $w$ converges to a specific driving culture.

At last, with $\Psi$ and $w$, Equation \ref{eq4} can now be transformed to:
\begin{equation}
Q^{\pi,w}(s, a) = \Psi^\pi(s, a)^\top\cdot w = \mathbb{E}_{\pi} \left[ \sum_{t=0}^{\infty} \gamma^t \Phi(s_t, a_t)^\top \cdot w_{t} \right]
\end{equation}
where $\Psi^\pi(s, a)^\top\cdot w$ denotes the inner product of the $\Psi^\pi(s, a)$ and $w$. Consequently, we can establish the action value function for open-world human driving without explicit rewards, thus understanding human driving behavioral patterns by combining the driving archetype with specific driving cultures.

\subsubsection*{Model Architecture}\label{sec422}
\begin{figure}[h]
    \centering
    \includegraphics[width=1\textwidth]{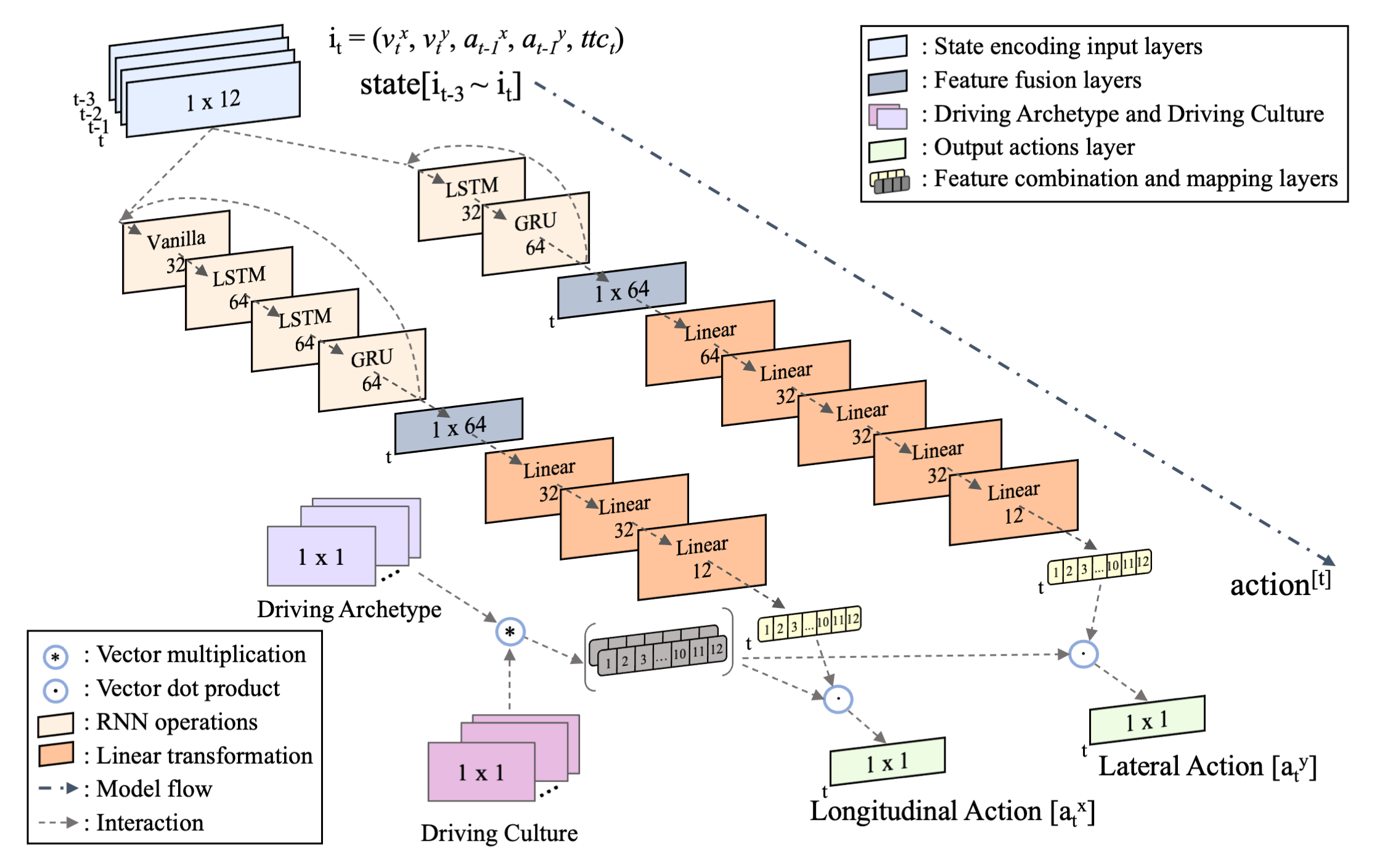}
    \caption{Neural network architecture for modeling driving behavior with temporal dynamics. The proposed model processes a sequence of state vectors $(i_{t-3}, i_{t-2}, i_{t-1}, i_t)$, capturing the ego vehicle’s velocity, acceleration, and TTC in eight directions, leveraging LSTM, GRU, and RNN modules. This architecture separately models lateral and longitudinal driving behaviors due to their distinct action-state distributions. The network encodes states through feature fusion layers and generates the successor feature and cultural vector for policy evaluation. Finally, the network outputs longitudinal and lateral actions, $(a_t^x, a_t^y)$, optimizing the mapping from states to actions via supervised learning and generalized policy iteration (GPI).}
    \label{fig5}
\end{figure}
Considering that human driving is continuous, we craft a neural network to capture the target $\Psi^\pi(s, a)$. As shown in Fig. \ref{fig5}, our model leverages LSTM, gated recurrent unit (GRU), and recurrent neural network (RNN) modules, which can feature driving behavior with temporal dynamics. Especially, due to the difference between the distributions of actions and states on lateral and longitudinal dimensions, we designed two separate network structures to model driving behaviors of different dimensions.

Our model takes the ego vehicle's velocity, acceleration, and TTC as input (i.e., state), and takes the acceleration as output (i.e., action). Specifically, for a vehicle, its $v$, $a$, and $ttc$ in eight directions at the current time step $t$ can be represented by a state vector $i_t = (v_t^x, v_t^y, a_{t-1}^x, a_{t-1}^y, \{ttc_t^i\}_{i=1}^8|t)$. Next, four history consecutive frames of state vectors, represented by $(i_{t-3}, i_{t-2}, i_{t-1}, i_t)$,  are considered as the input. As to the output, $a_{t}$ in two directions is defined as the output action, represented by $a_{t} = (a_{t}^x, a_{t}^y)$.

Training our network is based on supervised learning. The model receives the state before $t$ and outputs the action for $t$, with the action recorded in real-world driving data as the learning target. The loss function at $t$ for training is based on mean squared errors (MSE), comparing the difference between the model output and the target one, expressed by:
\begin{equation}
    \mathcal{L}(a) = \frac{1}{N} \sum_{i=1}^{N} (\hat{a_t} - a_t)^2
\end{equation}
where $\hat{a}$ and $a$ represent the model output action and the target action recorded in human driving data at $t$, respectively, and $N$ is the batch size (128 in this paper). By minimizing this loss, the model can optimize its weights to learn the underlying driving archetype. 

Here we employed the generalized policy iteration (GPI) algorithm to improve model performance. GPI alternates between evaluating the current policy and updating the policy to maximize expected returns, by which the model performance can be continuously enhanced. In this context, the policy can be regarded as the learned mapping from states to actions. Specifically, GPI encompasses two main steps: 1) policy evaluation, which evaluates the action-value function $Q^{\pi_k}(s, a)$ under the currently learned policy $\pi_k$; and 2) policy improvement, which updates the policy $\pi$ to maximize the action-value function $Q^{\pi}(s, a)$, thereby increasing the expected return for each state-action pair. The calculation of GPI can be expressed as:
\begin{equation}
    \pi_{k+1}(s) = \arg\max_{a} \left[ Q^{\pi_k}(s, a) + \gamma \sum_{s'} P(s'|s, a) \max_{a'} Q^{\pi_k}(s', a') \right]
\end{equation}
where $Q^{\pi_k}(s, a)$ is the action value function under the current policy $\pi_k$, and $\pi_{k+1}$ is the updated policy. $P(s'|s, a)$ is the probability of transitioning to state $s'$ after taking action $a$ at state $s$, reflecting the dynamics of state transitions. The $\arg\max$ indicates selecting the action $a$ that can maximize the action-value function $Q^{\pi_k}(s', a')$ for action $a'$ in state $s'$, where $a'$ means the possible actions at $s'$. The discount factor $\gamma$ is used to discount the future action value. Through this iterative process, the policy can be gradually updated until it converges to the optimal solution that aligns with human driving data.

\subsection*{Cross-cultural Deployment}\label{sec423}
In previous illustrations, we detail the algorithm and model architecture of DLIRL, where the extracted driving archetype $\Psi$ and driving culture $w$ have been formalized. In this section, we will describe how to operationalize cross-cultural deployment.

First, the driving archetype $\Psi$ needs to be extracted from a dataset specific to a particular driving culture. Specifically, we first initialize $w$ to an all-one vector to represent the current driving culture. Keeping $w$ fixed, $\Psi$ can be learned from the data to capture the driving archetype, as illustrated in the sections above. 

Next, we can transfer the extracted driving archetype across different driving cultures. Given another culture-specific dataset, we can transfer and reserve the previously learned driving archetype by keeping $\Psi$ fixed. Thus, the model can be trained to align $w$ with the new driving culture. Under the aegis of the shared driving archetype, the driving culture can be learned and captured at a lower data cost. In doing so cross-cultural deployment can be done efficiently and effectively.

%\section*{Acknowledgements}
%This study is supported by RGC General Research Fund (GRF) HKUST16205224.

%\section*{Author contributions statement}
%:
%A.A. conceived the experiment(s), H.L. and B.A. conducted the experiment(s), C.A. and D.A. analysed the results. All authors reviewed the manuscript. 

%\section*{Competing interests} 
%The authors declare no competing interests.

\newpage
\bibliography{sample}

%\section{Supplementary information}
% -介绍互训实验
% 在正文中我们阐释了将基于HighD文化的sf与cv跨文化应用的结果，并展示了运动分布、驾驶风格、驾驶安全的deployment结果。在本节我们将展示分别基于Dji与NGSIM文化的跨文化部署结果以及对比展示

% 实验结果说明
\end{document}